\documentclass[11pt,a4paper]{article}
\usepackage[hyperref]{emnlp2020}

\usepackage{times}
\usepackage{latexsym}
\usepackage{adjustbox}
\usepackage{multirow}
\usepackage{xspace}
\usepackage{booktabs}
\usepackage{tabularx}

\newcommand{\total}{\mathrm{total}}
\newcommand{\gen}{\mathrm{gen}}
\newcommand{\type}{\mathrm{type}}
\newcommand{\connec}{\mathrm{conn}}
\newcommand{\disco}{\textsc{DiscoFuse}}
\newcommand{\BertTrans}{\texttt{BERT}}
\newcommand{\AuxBertTrans}{\texttt{AuxBERT}}
\newcommand{\AugBertTrans}{\texttt{AugBERT}}
\newcommand{\AugAuxBertTrans}{\texttt{AugAuxBERT}}

\newcommand{\Trans}{\texttt{Transformer}}
\newcommand{\LaserTagger}{\texttt{LaserTagger}\xspace}
\newcommand{\MREXACT}{\textsc{MR-Exact}}
\newcommand{\MRSARI}{\textsc{MR-SARI}}
\newcommand{\EXACT}{\textsc{Exact}}
\newcommand{\SARI}{\textsc{SARI}}

\newcommand{\qex}[1]{`\emph{#1}'}
\usepackage{amsmath}
\usepackage{cleveref}
\usepackage{bbm}

\definecolor{darkgreen}{rgb}{0.1, 0.7, 0.1}
\definecolor{darkred}{rgb}{0.7, 0.1, 0.1}

\newcolumntype{L}[1]{>{\raggedright\arraybackslash}p{#1}}
\newcolumntype{C}[1]{>{\centering\arraybackslash}p{#1}}
\newcolumntype{R}[1]{>{\raggedleft\arraybackslash}p{#1}}

\usepackage{microtype}

\aclfinalcopy % Uncomment this line for the final submission
\def\aclpaperid{1288} %  Enter the acl Paper ID here

\title{Semantically Driven Sentence Fusion: Modeling and Evaluation}

\author{Eyal Ben-David\textsuperscript{1}, Orgad Keller\textsuperscript{2}, Eric Malmi\textsuperscript{2}, Idan Szpektor\textsuperscript{2}, Roi Reichart\textsuperscript{1} \\
  \textsuperscript{1}Faculty of Industrial Engineering and Management, Technion, IIT \\
  \textsuperscript{2}Google Research \\
  \texttt{eyalbd12@campus.technion.ac.il, roiri@technion.ac.il} \\
  \texttt{\{orgad,emalmi,szpektor\}@google.com} \\}

\date{}

\begin{document}
\maketitle
\begin{abstract}
Sentence fusion is the task of joining related sentences into coherent text.
Current training and evaluation schemes for this task are based on single reference ground-truths and do not account for valid fusion variants. We show that this hinders models from robustly 
capturing the semantic relationship between input sentences.
To alleviate this, we present an approach in which ground-truth solutions are automatically expanded into multiple references via curated equivalence classes of connective phrases. We apply this method to a large-scale dataset and use the augmented dataset for both model training and evaluation.
%In order to direct a fusion generation model to better capture the semantic relationship in the input as represented in the augmented training set, 
To improve the learning of semantic representation using multiple references, 
we enrich the model with auxiliary discourse classification tasks under a multi-tasking framework. Our experiments highlight the improvements of our approach over state-of-the-art models. 
\footnote{Our code is at \url{https://github.com/eyalbd2/Semantically-Driven-Sentence-Fusion}.}
% Un-comment the following footnote for arxiv version
\footnote{This paper was accepted to Findings of EMNLP 2020.}

\end{abstract}

\section{Introduction}
\label{sec:intro}

Generative NLP tasks, such as machine translation and summarization, often rely on human generated ground truth.
%, i.e., the target translation or summarization, for evaluation.  
Datasets for such tasks typically contain only a single reference per example. This may result from the costly effort of human annotations, or from collection methodologies that are restricted to single reference resources (e.g., utilizing existing corpora; \citealp{koehn2005europarl,nallapati-etal-2016-abstractive-fixed}). However, typically there are other possible generation results, such as ground-truth paraphrases, that are also valid. Failing to consider multiple references hurts the development of generative models, since such models are considered correct, at both training and evaluation, only if they follow one specific and often arbitrary generation path per example.

In this work we address \emph{Sentence Fusion}, a challenging task where a model should combine related sentences, which may overlap in content, into a compact coherent text. The output should preserve the information in the input sentences as well as their semantic relationship.
It is a crucial component in many NLP applications, including text summarization, question answering and retrieval-based dialogues 
\citep{jing2000cut, barzilay2005sentence, marsi2005explorations, DBLP:journals/corr/abs-1910-00203, szpektor2020dynamic}.

Our analysis of state-of-the-art fusion models \citep{geva2019discofuse, rothe2019leveraging} 
indicates that they still struggle to correctly detect the semantic relationship between the input sentences, which is reflected in inappropriate discourse marker selection in the generated fusions (\S \ref{sec:learning}). 
At the same time, \disco{} \citep{geva2019discofuse}, the dataset they use, is limited to a single reference per example, ignoring discourse marker synonyms such as `but' and `however'. Noticing that humans tend to judge these synonyms as equally suitable (\S \ref{sec:multi_ref}), we hypothesize that relying on single references may limit the performance of those models.

%take advantage of the recently introduced large-scale fusion dataset \disco{} \citep{geva2019discofuse} to 
%train large neural architectures that improve greatly over previous approaches. But, our analysis
%alternative phrasings, e.g. `but' and `however', are not captured by the single gold-label reference point offered in the dataset,
%\EXACT{} and \SARI{} \citep{SARI}, 
%while humans tend not to differentiate between them \citep{geva2019discofuse}.
%When the semantics of the discourse relation is not emphasized by the loss function and is not accounted for during evaluation, it is not surprising that it is not properly captured in current fusion models.

%We conjecture that the reason for the incorrect detection of the semantic relation is the lack of semantic directive from the standard token-level cross-entropy-based loss employed at training time. Specifically, seq2seq fusion models need to learn to copy much of their input to their output. The cross-entropy loss accentuates the frequent copy mistakes at the expense of sparser semantic errors, which are reflected by few word changes, such as replacing `but' with `also'. 

To overcome this limitation, we explore an approach in which ground-truth solutions are automatically expanded into multiple references.
Concretely, connective terms in gold fusions are replaced with equivalent terms (e.g., \{\qex{however}, \qex{but}\}
%and  \{\qex{moreover}, \qex{additionally}\}
), where the semantically equivalent sets are derived from the Penn Discourse TreeBank 2.0 \citep{DBLP:conf/lrec/PrasadDLMRJW08}. Human evaluation of a sample of these generated references indicates the high quality of this process (\S \ref{sec:multi_ref}).
%
%Given the justifications to our expansion process, both from theory and from  human-judgement, we believe that our multi- reference dataset can be reliably used both for evaluation and for model training. 
We apply our method to automatically augment the \disco{} dataset with multiple references, using the new dataset both for evaluation and training.
%As a first contribution of this paper 
We will make this dataset publicly available.

We then adapt a seq2seq fusion model in two ways so that it can exploit the multiple references in the new dataset (\S \ref{sec:learning}).
First, each training example with its multiple references is converted into multiple examples, each consisting of the input sentence pair with a different single reference fusion. Hence, the model is exposed to a more diverse and balanced set of fusion examples.
%Therefore, while the input to the model consists of single-reference solutions, it is exposed to a more diverse and balanced set of fusion examples. 
Second, we direct the model to learn a common semantic representation for equivalent surface forms offered by the multiple references. To that end, we enhance the model with two auxiliary tasks: Predicting the type of the discourse relation and predicting the connective pertaining to the fused output, as derived from the reference augmentation process.

We evaluate our model against state-of-the-art models in two experimental settings (\S \ref{sec:settings}, \ref{sec:results}): In-domain and cross-domain learning. The cross-domain setting is more challenging but may also be more realistic as labeled data is available only for the source domain but not for the target domain. 
To evaluate against multiple-reference examples, we measure the similarity of each generated fusion to each of the ground-truth fusions and report the highest score. This offers a more robust analysis, and reveals that the performance of fusion models is higher than previously estimated.
In both settings, our model demonstrates substantial performance improvement over the baselines.

% We evaluate our model against multiple-reference examples, measuring the similarity of the generated fusion to each of the ground-truth fusions and reporting the highest score.
% %(MR \SARI{} or MR \EXACT{}). 
% This offers a more robust analysis, and can reveal that the performance of a model is higher than previously estimated. We note that this evaluation approach is complementary to recall-oriented evaluation approaches for single references that employ semantic similarity measures instead of exact word matching \citep{wieting2019beyond}. While the recall-oriented approaches may detect fusion outputs that are beyond the reach of our approach as correct, they may also promote erroneous solutions due to their soft matching nature.

% We apply this evaluation in two experimental settings (\S \ref{sec:settings}, \ref{sec:results}): In-domain and cross-domain learning. The cross-domain setting is more challenging but may also be more realistic as labeled data is available only for the source domain but not for the target domain. In both setups, our model demonstrates substantial performance gains over a variety of baselines, including state-of-the-art models\footnote{Our code will be made public upon paper acceptance.}.

\section{Related Work}
\label{sec:related_work}

\subsection{Fusion Tasks}

%Until recently, supervised sentence fusion models had access to only small labeled datasets. Therefore, they relied on hand-crafted features stemming from dependency syntax \citep{barzilay2005sentence, filippova2008sentence, elsner2011learning} or adjacency relationships in a word graph \cite{filippova2010multi}, or on uni-gram and bi-gram features employed in structured prediction models \cite{thadani2013supervised}.

Traditionally, supervised sentence fusion models had access to only small labeled datasets. Therefore, they relied on hand-crafted features \citep{barzilay2005sentence, filippova2008sentence, elsner2011learning,filippova2010multi,thadani2013supervised}.
Recently, \disco{}, a large-scale dataset for sentence fusion, was introduced by \citet{geva2019discofuse}. 
%To avoid the laborious process of manual example preparation, 
This dataset was generated by automatically applying hand-crafted rules for 12 different discourse phenomena to break fused text examples from two domains, Wikipedia and Sports news, into two un-fused sentences, while the content of the original text is preserved. We follow prior work \citep{malmi2019encode,rothe2019leveraging} and use the balanced version of \disco{}, containing $\sim$16.5 million examples, where the most frequent discourse phenomena were down-sampled. 
%The dataset contains $\sim$16.5 million examples from two domains: Wikipedia and Sports. 
%Each example is augmented with structured information about the discourse phenomenon it expresses, and the connective term that was used in the original fused sentence.

With \disco{}, it became possible to train data-hungry neural fusion models.
% can now be employed to the task, and, perhaps unsurprisingly, they achieve state-of-the-art results. % Eric: AFAIK, NN approaches and the pre-NN approaches haven't been compared.
\citet{geva2019discofuse} showed that a Transformer model \citep{vaswani2017attention} outperforms an LSTM-based \citep{LSTM} seq2seq model on this dataset. \citet{malmi2019encode} further improved accuracy by introducing LaserTagger, modeling sentence fusion as a sequence tagging problem. \citet{rothe2019leveraging} set the state-of-the-art with a BERT-based \cite{devlin2018bert} model.
%, which serves as our baseline.

%Related to sentence fusion is the task of predicting the discourse marker that should connect two input sentences. \citet{elhadad1990generating} and \citet{grote1998discourse} developed hand-crafted features for distinguishing between connectives.  \citet{malmi2017automatic} applied a decomposable attention model to this task.
%, observing that the model is able to predict the ground-truth connective more often than human raters.

Related to sentence fusion is the task of predicting the discourse marker that should connect two input sentences \citep{elhadad1990generating,grote1998discourse,malmi2017automatic}.
It is typically utilized as an intermediate step to improve downstream tasks, mainly for discourse relation prediction \citep{pitler2008easily, zhou2010effects,braud2016learning,qin2017adversarial}. %liu2016recognizing, qin2016stacking
Connective prediction was included in multi-task frameworks for discourse relation prediction
\citep{liu2016implicit} and unsupervised sentence embedding \citep{jernite2017discourse,nie2019dissent}. We follow this approach of guiding a main task with the semantic information encompassed in discourse markers, studying it in the context of sentence fusion.

\subsection{Generation Evaluation}
\label{ssec:related_eval}

Two main approaches are used to evaluate generation models against a single gold-truth reference. The first estimates the correctness of a generated text using a `softer' similarity metric between the text and the reference instead of exact matching.
Earlier metrics like BLEU and ROUGE \citep{papineni-etal-2002-bleu,lin-2004-rouge}, considered n-gram agreement. Later metrics matched words in the two texts using their word embeddings \citep{lo-2017-meant,clark-etal-2019-sentence}. More recently, contextual similarity measures were devised for this purpose \citep{lo-2019-yisi,wieting2019beyond,zhao-etal-2019-moverscore,bert-score,sellam2020bleurt}. In \S \ref{sec:qual} we provide a qualitative analysis for the latter, presenting typical evaluation mistakes made by a recently-proposed contextual-similarity based metric \citep{bert-score}. This analysis reveals properties that characterize such methods, which make them less suitable for our task.

The second approach extends the (single) reference into multiple ones, by automatically generating paraphrases of the reference (a.k.a pseudo-references)
%, which provide more variability in both training and testing 
\citep{albrecht-hwa-2008-role,yoshimura-etal-2019-filtering,kauchak-barzilay-2006-paraphrasing,edunov-etal-2018-classical,gao2020supert}.
Our method (\S \ref{ssec:eval_variants}) follows this paradigm. It applies curated paraphrase rules to generate highly accurate variations, putting an emphasis on precision.
This is opposite to the recall-oriented similarity-based approaches, which may detect correct fusions beyond paraphrasing approaches, but may also promote erroneous solutions due to their soft matching nature.

\begin{table*}[t]
\centering

%% local settings
\small
% for avoiding siunitx using bold extended
\renewrobustcmd{\bfseries}{\fontseries{b}\selectfont}
\renewrobustcmd{\boldmath}{}
% abbreviation
\newrobustcmd{\B}{\bfseries}

\begin{tabular}{c  p{13.2cm}}
 \hline
\textbf{Mistake type} & \textbf{Examples} \\
    \hline
     \multirow{3}{6em}{Correct fusion variant}  & \textcolor{violet}{(a+b)} It is situated around Bad Segeberg but not part of it . Bad Segeberg is the seat of the AMT .\\
                 & \textcolor{darkred}{(I)} It is situated around Bad Segeberg \textbf{,} the seat of the AMT \textbf{,} but not part of it .  \\
                 & \textcolor{darkgreen}{(G)} It is situated around Bad Segeberg \textbf{, which is} the seat of the AMT \textbf{,} but not part of it . \\
    \hline
     \multirow{3}{6em}{Missing/added anaphora} & \textcolor{violet}{(a+b)} Of the three , purple is preferred . Purple reinforces the red .\\
              & \textcolor{darkred}{(I)} Of the three , purple is preferred \textbf{because} purple reinforces the red .\\
              & \textcolor{darkgreen}{(G)} Of the three , purple is preferred \textbf{because it} reinforces the red .\\
    \hline
     \multirow{3}{6em}{Missing context/info}      & \textcolor{violet}{(a+b)} Bolger quickly defeated Mclay . Gair himself took the position of deputy leader .\\
                  & \textcolor{darkred}{(I)} Bolger quickly defeated Mclay \textbf{, while} Gair himself took the position of deputy leader . \\
                  & \textcolor{darkgreen}{(G)} Bolger quickly defeated Mclay \textbf{, and} Gair himself took the position of deputy leader . \\
    \hline
     \multirow{3}{6em}{Missing/added punctuation}   & \textcolor{violet}{(a+b)}  Gair was born in Dunedin . Gair was moved to Wellington when young . \\
                & \textcolor{darkred}{(I)} Gair was born in Dunedin \textbf{but} moved to Wellington when young . \\
                & \textcolor{darkgreen}{(G)} Gair was born in Dunedin \textbf{, but} moved to Wellington when young . \\
    \hline
     \multirow{3}{6em}{Annotation error}       & \textcolor{violet}{(a+b)} Krishnamurti 's notebook . By Krishnamurti , Krishnamurti ( hardcover ) .\\
                & \textcolor{darkred}{(I)} Krishnamurti ' s notebook . By krishnamurti ( hardcover ) . \\
                & \textcolor{darkgreen}{(G)} Krishnamurti ' s notebook . By Jiddu , krishnamurti ( hardcover ) . \\
    \hline
     \multirow{3}{6em}{Semantic Errors}     & \textcolor{violet}{(a+b)} George Every never married . George Every never had children.  \\
                  & \textcolor{darkred}{(I)} George Every never married \textbf{or} had children .  \\
                  & \textcolor{darkgreen}{(G)} George Every never married \textbf{nor} had children . \\
    \hline

\end{tabular}
\caption{Examples of various model error types. The input text is marked with \textcolor{violet}{a+b}, the generated fusion is marked with \textcolor{darkred}{I} and the ground-truth fusion is marked with \textcolor{darkgreen}{G}. Errors are highlighted in bold font.}
\label{tab: examples of model mistakes.}
\end{table*}

\begin{table}[t]
\centering
\small
\begin{tabular}{l|r|r}
Error Type & Wikipedia & Sports \\
\hline
Correct Fusion Variant & 40\% & 44\% \\
Miss/Add Anaphora & 2\% & 2\% \\
Missing Context & 18\% & 18\% \\
Miss/Add Punctuation & 22\% &  18\% \\
Annotation Error & 8\% & 8\% \\
Semantic Error & 10\% & 10\% \\
\end{tabular}
% \begin{tabular}{ c | c | c | c | c | c | c}
% & {Correct} & {Miss/Add} & {Missing} & {Miss/Add} & {Annotation} & {Semantic} \\
% & {Fusion Variant} & {Anaphora} & {context} & {Punctuation} & {error} & {error} \\
%     \hline
%      Wikipedia & 40\% & 2\% & 18\% & 22\% & 8\% & 10\% \\
%      \hline
%      Sports  & 44\% & 2\% & 18\% & 18\% & 8\% & 10\% \\
% \end{tabular}
\caption{Error type distribution in 50 dev examples.} 
\label{tab:Mistakes-statistics}
\end{table}
%\eric{Should we move these numbers to Table 1 to save space?}

\begin{table}[t]
\centering
\small
% \begin{adjustbox}{width=0.4\textwidth}
\begin{tabular}{ @{}p{1.8cm} | p{2.1cm} | l@{} }
% \textbf{Contingency\newline$\rightarrow$ Cause} & \textbf{Expansion\newline$\rightarrow$ Conjunction} & \textbf{Comparison}   \\
\textbf{Cause} & \textbf{Conjunction} & \textbf{Comparison}   \\
\hline
 As a result & Furthermore & However   \\
%  \hline
Hence & And & Yet \\
%  \hline
Consequently & Additionally & Still \\
%  \hline
Thus & Moreover & Nevertheless \\
%  \hline
Therefore & Plus & Although \\
 &  & But \\
\end{tabular}
% \end{adjustbox}
\caption{Clusters of interchangeable connective markers constructed based on PDTB 2.0 sense tags.
%Notice that this is not the exact way that these words may appear in text. For example, 'hence' might appear in the middle of a sentence (with a comma before it), and it can also appear at the beginning of a sentence. 
}
\label{tab:Clustering-connectives}
\end{table}

\section{Multiple References in Sentence Fusion}\label{sec:multi_ref}
In this section we discuss the limitations of using single references for evaluation and training in sentence fusion. We then propose an automatic, precision-oriented method to create valid fusion variants. Human-based evaluation confirms the reliability of our method, which generates pseudo-references that are considered as suitable as the original references. Finally, we demonstrate the effectiveness of the new references for fusion evaluation. Our observations, which are used here for reference generation and evaluation, will also guide our fusion model design and training (\S \ref{sec:learning}).
% In this section we discuss the limitations of standard fusion evaluation metrics to detect valid fusion variants of a given gold label fusion. We propose a new evaluation metric that addresses this issue. Our observations, which are used here to improve evaluation, will guide our novel modeling design and training procedure in \Cref{sec:learning}. 

%In this section we discuss the various categories of fusion cases, observing that some categories can be realized by different but semantically equivalent discourse markers. We further observe that the standard evaluation metrics of the task cannot properly account for such cases, and propose a new evaluation framework that reduces this gap. Our observations, which are used here for better evaluation, will guide our novel modeling design and training procedure in Section \ref{sec:learning}.

\begin{table*}[t]
\centering

%% local settings
% \begin{adjustbox}{width=\textwidth}
\small
% for avoiding siunitx using bold extended
\renewrobustcmd{\bfseries}{\fontseries{b}\selectfont}
\renewrobustcmd{\boldmath}{}
% abbreviation
\newrobustcmd{\B}{\bfseries}

\begin{tabular}{ l c p{12.5cm}}
 \hline
\textbf{Phenomenon}  & &  \textbf{Examples}\\
    \hline
     Conjunction  & \textcolor{darkgreen}{G} & It'll work because god says so . \textbf{Plus ,} we are both willing to fight for it . \\
        & \textcolor{blue}{V} & It'll work because god says so . \textbf{Furthermore ,} we are both willing to fight for it .\\
        & \textcolor{blue}{V} & It'll work because god says so \textbf{, and} we are both willing to fight for it .\\
        
    \hline
     Cause  & \textcolor{darkgreen}{G} & But the client is on a break. \textbf{Therefore} I'm on a break.\\
        & \textcolor{blue}{V} & But the client is on a break. \textbf{Hence} I'm on a break. \\
        
    \hline
     Comparison  & \textcolor{darkgreen}{G} & It might sound like a nightmare \textbf{but} this news made this day one of the greatest of my life . \\
    % & \textcolor{blue}{V} & It might sound like a nightmare \textbf{although} this news made this day one of the greatest of my life . \\
    & \textcolor{blue}{V} & It might sound like a nightmare \textbf{. Yet ,}  this news made this day one of the greatest of my life . \\
     
    \hline
    \multirow{2}{1.5CM}{Relative Clause}   & \textcolor{darkgreen}{G} & She is famed for her noble art Raikiri, \textbf{which is a} slash powered by lightning, that is believed to be inevitable. \\
     & \textcolor{blue}{V} & She is famed for her noble art Raikiri, \textbf{a} slash powered by lightning, that is believed to be inevitable. \\
    %  Relative   & \textcolor{red}{a} & Jonouchi has a crush on Miyuki , \textbf{who is} the nurse taking care of his sister , Shizuka . \\
    % Clause & \textcolor{blue}{a} & Jonouchi has a crush on Miyuki , the nurse taking care of his sister , Shizuka . \\
    \hline
\end{tabular}
% \end{adjustbox}
\caption{Examples of automatic variant generation for fusion phenomena. The gold fusion is marked by \textcolor{darkgreen}{G}. The automatically generated variants are marked by \textcolor{blue}{V}. Parts that were changed during variant generation are \textbf{boldfaced}.}
\label{tab: examples of data aumentation.}
\end{table*}

\subsection{Single-reference Based Evaluation}
\label{ssec:eval_metric_analysis}

Recent fusion works \citep{geva2019discofuse, malmi2019encode, rothe2019leveraging} rely on single references for training and evaluation. Two evaluation metrics are used: (1) \EXACT, where the generated fusion should match the reference exactly, and (2) \SARI{} \citep{SARI}, which measures the F1 score of kept and added n-grams, and the precision of deleted n-grams, compared to the gold fusion and the input sentences, weighting each equally.

A significant limitation of the above metrics, when measured using a single fusion reference, is that they do not properly handle semantically equivalent variants. 
For \EXACT{} this is obvious, since even one word difference would account as an error.
%For example, in \EXACT{}, a fusion where the connective marker is different than the one in the ground-truth fusion would count as an error, even if the connectives are semantically similar. 
In \SARI{}, 
%while a single word difference would not be counted as a full error, 
the penalty for equivalent words, e.g., \qex{but} and \qex{however}, and non-equivalent ones, e.g., \qex{but} and \qex{moreover}, is identical.   

To validate this, we conducted a qualitative single-reference evaluation of a fusion model (\AuxBertTrans, \S \ref{sec:auxiliary model}) under the \EXACT{} metric.
We randomly selected 50 examples, assessed as mistakes, from the dev sets of both \disco{}'s domains. 
Analyzing these mistakes, we identified six types of errors (\Cref{tab: examples of model mistakes.}).
% \eric{It's not clear which model has produced these incorrect fusions.} 

The distribution of these error types is depicted in \Cref{tab:Mistakes-statistics}.
We note that the most frequent error type refers to \emph{valid} fusion variants that differ from the gold fusion: As much as 40\% and 44\% of the examples in the Wikipedia and the Sports datasets, respectively. 
%The second most frequent error type is syntactic in nature -- punctuation mistakes.
While the sample size is too small for establishing accurate statistics, the high-level trend is clear, indicating that a significant portion of the generated fusions classified as mistakes by the \EXACT{} metric are in fact correct.

A possible solution would be to rely on single references, but use `softer' evaluation metrics (see \S \ref{ssec:related_eval}).
We experimented with the state-of-the-art BERTScore metric \citep{bert-score} and noticed that it often fails to correctly account for the semantics of discourse markers (see \S \ref{sec:qual}), which is particularly important for sentence fusion. 
Furthermore, we notice that recent soft metrics depend on trainable models, mainly BERT \citep{devlin2018bert}, which is also used in state-of-the-art fusion models \citep{malmi2019encode, rothe2019leveraging}. Thus, we expect these metrics to struggle in evaluation when fusion models struggle in prediction.
%However, we notice two  limitations in applying such metrics for sentence fusion. First, soft metrics depend on trainable models that are often in use within fusion models. Thus, we expect these metrics to struggle in evaluation when fusion models struggle in prediction. \orgad{I think we should emphasize the following part much more: ``we have tested the applicability of a SotA metric, BERTScore, for evaluating fusion. In many cases, we see that the metric given a similar score to...''} Second, as demonstrated in the supplementary material, these metrics fail to correctly account for the semantics of discourse markers, which is particularly important for sentence fusion.

\subsection{Multi-Reference Generation}
\label{ssec:fusion_variants}

Generation of valid variants that differ from the ground-truth reference is a challenge for various generation tasks. For open-ended tasks like text summarization, researchers often resort to manually annotating multiple valid reference summaries for a small sample of examples. Sentence fusion, however, is a more restricted task, enabling high-quality automatic paraphrasing of gold fusions into multiple valid references. We introduce a precision-oriented approach for this aim.

%Our method is focused on creating semantically equivalent variants of ground-truth fusions. To this end, 

Instead of generating arbitrary semantically equivalent paraphrases, we focus on generating variants that differ only by discourse markers, which are key phrases to be added when fusing sentences. The \emph{Penn Discourse TreeBank 2.0} (PDTB; \citealp{DBLP:conf/lrec/PrasadDLMRJW08}) contains examples of argument pairs with an explicit discourse marker and a human-annotated sense tag. The same marker may be associated with multiple sense tags (for instance, \textit{since} may indicate both temporal and causal relations), and for our precision-oriented approach we only considered unambiguous markers. 

%Concretely, we identified three tags whose markers cover more than 90\% of all argument pairs in the DiscoFuse dataset (see the header line of \Cref{tab:Clustering-connectives}).
% These PDTB tags induce equivalence clusters, presented in \Cref{tab:Clustering-connectives}, which include the markers related to each tag .
Concretely, we identified three PDTB sense tags most relevant to the DiscoFuse dataset and chose the markers whose tag is one of those three in at least 90\% of all PDTB argument pairs with an explicit marker. The resulting clusters are presented in \Cref{tab:Clustering-connectives}.\footnote{Some connective occurrences differ only in the addition or omission of a punctuation mark, e.g., \qex{but} and \qex{but,}. From a sample of examples, we did not find cases in which punctuation changes the semantics of the resulting connection. Therefore, for every connective phrase in \Cref{tab:Clustering-connectives}, we automatically consider also its variants that differ only in punctuation. 
%We hence cluster 29 discourse markers out of the 71.
} % Finally, we add a fourth cluster containing the phrases \textit{who is the} and \textit{which is a} to cover relative clause paraphrases.
Finally, we add a fourth cluster containing relative clause paraphrases (such as \textit{who is the} and \textit{which is a}). Paraphrases from this cluster are not equivalent and cannot be replaced one with each other. Instead, they are replaceable with apposition paraphrases (as demonstrated in \Cref{tab: examples of data aumentation.}, under \textit{Relative Clause}).

% Our method follows the Penn TreeBank annotations of \emph{Implicit Connectives} \citep{DBLP:conf/lrec/PrasadDLMRJW08} to construct semantically equivalent clusters of discourse markers. We then manually analyze the connective phrases (a.k.a connective discourse markers) appearing in \disco{} and associate them to the appropriate cluster, as presented in \Cref{tab:Clustering-connectives}. This results with four semantic clusters: 
%  \emph{Addition}, \emph{Effect}, \emph{Qualify} and \emph{Relative Clause}.\footnote{Some of the connective phrases in \disco{} differ only in the addition or omission of a punctuation mark (e.g. \emph{but} and \emph{but,}). Every connective phrase in the table also refers to its variants that differ only in punctuation, because punctuation does not change the semantics of the resulting connection. We hence cluster 29 discourse markers out of the 71.} 

Given a \disco{} target fusion $t_i$, if $t_i$ is annotated with a connective phrase $c_i$ that appears in one of our semantic clusters, we define the set $\mathcal{V}(t_i)$ that includes $t_i$ and its variants. These variants are automatically generated by replacing the occurrence of $c_i$ in $t_i$ by the other cluster members. \Cref{tab: examples of data aumentation.} demonstrates this process. More details and examples are in the appendix (\S \ref{sec:apend-augmentation-rules}).

\subsection{The Quality of Multiple References}
\label{ssec:eval_variants}

To validate the reliability of our automatically generated variants as ground-truth fusions, we evaluate their quality with human annotators. To this end, we randomly sampled 350 examples from the \disco{} dev sets (Wikipedia and Sports). Each example consists of two input sentences, and two fusion outcomes: the gold fusion and one automatically generated variant. We then conducted---using Amazon Mechanical Turk---a crowd-sourcing experiment where each example was rated by 5 native English speakers. Each rater indicated if one fusion outcome
%, the original gold fusion or the automatically generated one, 
is better than the other, or if both outcomes are of similar quality (good or bad). We considered the majority of rater votes for each example.
\Cref{tab:augmentation eval} summarizes this experiment. It shows that the raters did not favor a specific fusion outcome, which reinforces our confidence in our precision-based automatic generation method. 
%Furthermore, when asked later about their overall ratings, the raters often indicated that they preferred the automatic variant over the gold, although both were correct. We note that, as expected, whenever the original gold fusion was evaluated to be of bad quality, the automatic variant was also rated as bad.

To demonstrate the benefit of our generated multiple references in fusion evaluation, we re-evaluated the mistakes marked by single-reference \EXACT{} in \S \ref{ssec:eval_metric_analysis}. Concretely, each gold fusion $t_i$ was automatically expanded into the multiple reference set $\mathcal{V}(t_i)$.
We define a multi-reference accuracy: \MREXACT{} $= 1/N \sum_{i=1}^N \max_{t \in \mathcal{V}(t_i)}\mathbbm{1}[f_i=t]$, 
where $f_i$ is the generated fusion for example $i$, $\mathbbm{1}$ is the indicator function, and $N$ is the size of the test-set.
\MREXACT{}\footnote{We also define the \MRSARI{} measure. It follows \MREXACT's formulation, taking the maximum over the \SARI{} score between the generated fusion and the references.} considers a generated fusion for an example correct if it matches one of the variants in $\mathcal{V}(t_i)$.
% We then computed the max over all reference comparisons to the generated fusion $f_i$: $\MRExact = \max[EXACT(f_i, t) | t \in \mathcal{V}(t_i)]$.
We measured an absolute error reduction of 15\% in both domains, where all these cases come from the \emph{correct fusion} class of \Cref{tab:Mistakes-statistics}.

%To demonstrate the benefit in using our $\mathcal{V}$ implementation on sentence fusion evaluation metrics, we re-evaluated the mistakes detected by single-reference \EXACT{} in the error analysis of \S \ref{ssec:eval_metric_analysis}, this time using the multi references. That is, each of the gold fusions, $t_i$, was automatically expanded into $\mathcal{V}(t_i)$, which was then compared to the generated fusion, taking the max score across all variants. Our variants reduced 15\% of the errors that were typed as \emph{correct fusion} in both domains (see \Cref{tab:Mistakes-statistics}), while not affecting any other error type. 

\begin{table}[t]
\centering
\small
\begin{tabular}{l|r}
\textbf{Rating Type} & \textbf{Preference (\%)} \\
\hline
Both equally good & 74.6 \\
Original fusion is better & 7.1 \\
Generated variant is better & 9.4 \\
Both equally bad & 2.3 \\
No majority & 6.6 \\
\end{tabular}
% \begin{tabular}{c|c|c|c|c}
% & \textbf{Aug (\%)} & \textbf{Src (\%)} & \textbf{Both (\%)} & \textbf{Neither (\%)}  \\
% \hline
%  \textbf{Pref} & 17.5 & 17.5 & 56.0 & 10  \\
% \end{tabular}
\caption{Raters' preferences when comparing original \disco{} fusions to fusions generated by our automatic augmentation process.
%Examples are sampled from both the Wikipedia and the Sports domains.
}
% \eric{Are these numbers for Wikipedia or Sports?}\eyal{sampled half from wiki and half from sports.}}
\label{tab:augmentation eval}
\end{table}

\begin{table*}
\begin{center}
\small
\begin{tabular}{ c l }
 \hline
& \textbf{Examples} \\
    \hline
    %  \textcolor{darkred}{(I)} & He served as president of Georgetown University \textbf{, but} decided to open the catholic institution to non-catholic students.\\
    %  \textcolor{darkgreen}{(G)} & He served as president of Georgetown University \textbf{and} decided to open the catholic institution to non-catholic students.\\
    % \hline
     \textcolor{darkred}{(I)} & Grace is told she can not get pregnant \textbf{and} IVF is unlikely to help.\\
     \textcolor{darkgreen}{(G)} & Grace is told she can not get pregnant \textbf{because} IVF is unlikely to help.\\
    \hline
    \textcolor{darkred}{(I)} & The mounds now appear smaller than they did in the past \textbf{because} extensive flooding in the centuries since their\\ 
    & construction has deposited 3 feet.\\
    \textcolor{darkgreen}{(G)} & The mounds now appear smaller than they did in the past \textbf{, although} extensive flooding in the centuries since their\\ 
    & construction has deposited 3 feet.\\
    \hline
    \textcolor{darkred}{(I)} & A Grand Compounder was a degree candidate at the University of Oxford who was required to pay extra for his\\
    & degree \textbf{because} he had a certain high level of income.\\
   \textcolor{darkgreen}{(G)} & A Grand Compounder was a degree candidate at the University of Oxford who was required to pay extra for his\\
   & degree \textbf{unless} he had a certain high level of income.\\
    \hline
    \textcolor{darkred}{(I)} & The Battalion lost 41 men killed \textbf{or} died of wounds received on 1 July 1916.\\
    \textcolor{darkgreen}{(G)} & The Battalion lost 41 men killed \textbf{and} died of wounds received on 1 July 1916.\\
    \hline
    %  \textcolor{darkred}{(I)} & Other communities persist in the old practice, since there is a teaching that states that \textbf{although} the underlying\\
    %  & reason behind a certain ruling has been cancelled.\\
    % \textcolor{darkgreen}{(G)} & Other communities persist in the old practice, since there is a teaching that states that \textbf{unless} the underlying reason\\
    % & behind a certain ruling has been cancelled.\\
    % \hline

\end{tabular}
\caption{Examples of semantic errors made by the \BertTrans{} model. The generated fusion is marked with \textcolor{darkred}{I} and the ground-truth fusion is marked with \textcolor{darkgreen}{G}. These examples are handled correctly by our \AuxBertTrans{} model.}
\label{tab: BERT mistakes.}
\end{center}
\end{table*}

\section{A Semantically Directed Fusion Model}
\label{sec:learning}

In the previous section we have established the importance of multiple references for sentence fusion. We next  show (\S \ref{sec:bert-lim}) that the state-of-the-art fusion model fails to detect the semantic relationship between the input sentences. We aim to solve this problem by expanding the training data to include multiple-references (MRs) per input example, where together these references provide a good coverage of the semantic relationship and are not limited to a single connective phrase. We then present our model (\S \ref{sec:auxiliary model}), which utilizes auxiliary tasks in order to facilitate the learning of the semantic relationships from the MR training data (\S \ref{sec:augmentation}).

\subsection{The SotA Model: Error Analysis}\label{sec:bert-lim}

\citet{rothe2019leveraging} set the current state-of-the-art results on the \disco{} dataset with a model that consists of a pre-trained BERT encoder paired with a randomly initialized Transfomer decoder, which are then fine-tuned for the fusion task. We re-implemented this model, denoted here by \BertTrans{}, which serves as our baseline.
We then evaluated \BertTrans{} on \disco{} using \MREXACT{} (\S \ref{ssec:eval_variants}) and report its performance on each of the discourse phenomena manifested in the dataset (\Cref{tab:sliced}). 

We found that this model excels in fusion cases that are \emph{entity-centric} in nature. In these cases, the fused elements are different information pieces related to a specific entity, such as pronominalization and apposition (bottom part of \Cref{tab:sliced}). These fusion cases do not revolve around the semantic relationship between the two sentences. This is in line with recent work that has shown that the pre-trained BERT captures the syntactic structure of its input text \cite{tenney-2019-bert}. 

On the other hand, the performance of the \BertTrans{} model degrades when fusion requires the detection of the \emph{semantic relationship} between the input sentences, which is usually reflected via an insertion of a discourse marker. Indeed, this model often fails to identify the correct discourse marker (top part of \Cref{tab:sliced}). \Cref{tab: BERT mistakes.} demonstrates some of the semantic errors made by \BertTrans{}.

\subsection{Automatic Dataset Augmentation}\label{sec:augmentation}

We aim to expose a fusion model to various manifestations of the semantic relation between the input sentences in each training example, rather than to a single one, as well as to reduce the skewness in connective occurrences. 
We hypothesize that this should help the model better capture the semantic relationship between input sentences. 

To this end, we utilize our implementation of the variant set $\mathcal{V}$ (\S \ref{ssec:fusion_variants}). Specifically, for each training example $(s_i^1,s_i^2,t_i)$, we include the instances $\{(s_i^1,s_i^2,t')\;|\;t' \in \mathcal{V}(t_i)\}$ to the augmented training set. We then train a fusion model on this augmented dataset. The augmented dataset balances between variants of the same fusion phenomenon because if in the original dataset one variant was prominent, its occurrences are now augmented with occurrences of all other variants that can be offered by $\mathcal{V}$. 
We denote the baseline model trained on the augmented dataset by \AugBertTrans{}.

\subsection{Semantically Directed Modeling}\label{sec:auxiliary model}

Multiple references introduce diversity to the training set that could guide a model towards a more robust semantic representation. Yet, we expect that more semantic directives would be needed to utilize this data appropriately. Specifically, we hypothesize that the lower performance of the state-of-the-art \BertTrans{} on semantic phenomena is partly due to its \emph{mean cross-entropy} (MCE) loss function:
\begin{equation*} \label{eq:2}
\ell_{\gen} = -\frac{1}{N}\sum_{i=1}^{N}\frac{1}{T_i} \sum_{j=1}^{T_i} \log p(t_{i,j} | s^1_i,s^2_i, t_{i,1..j-1})
\end{equation*}
where $N$ is the size of the training-set, $T_i$ is the length of the target fusion $t_i$,  $t_{i,j}$ is the $j$-th token of $t_i$, and $p(w | s^1_i,s^2_i,\mathrm{pre})$ is the model's probability for the next token to be $w$, given the input sentences $s^1_i,s^2_i$ and the previously generated prefix $\mathrm{pre}$. 

As discussed earlier, the word-level overlap between the fusion and its input sentences is often high. Hence, many token-level predictions made by the model are mere copies of the input, and should be relatively easy to generate compared to new words that do not appear in the input. However, as the MCE loss does not distinguish copied words from newly generated ones, it would incur only a small penalty if only one or two words in a long fused sentence are incorrect, even if these words form an erroneous discourse marker.
Moreover, the loss function does not directly account for the semantic (dis)similarities between connective terms. This may misguide the model to differentiate between similar connective terms, such as \qex{moreover} and \qex{additionally}.
%, or prefer one on top of the other. 

To address these problems, we introduce a multi-task framework, where the main fusion task is jointly learned with two auxiliary classification tasks, whose goal is to make the model explicitly consider the semantic choices required for correct fusion. 
The first auxiliary task predicts the type of discourse phenomenon that constitutes the fusion act out of 12 possible tags (e.g. \emph{apposition} or \emph{discourse connective}; see \Cref{tab:sliced}).
The second auxiliary task predicts the correct connective phrase (e.g. \qex{however}, \qex{plus} or \qex{hence}) out of a list of 71 connective phrases. As gold labels for these tasks we utilize the structured information provided for each \disco{} example, which includes the ground-truth discourse phenomenon and the connective phrase that was removed as part of the example construction.
We denote this model \AuxBertTrans{} and our full model with auxiliary tasks trained on the multiple-reference dataset \AugAuxBertTrans{}.

% The joint objective mimics a top-down generative process: first deciding on the high-level semantics of the output and on the concrete semantic phrasing that realizes it, and only then generating the fused text based on these decisions.

\begin{figure}[t]
\includegraphics[scale=0.45]{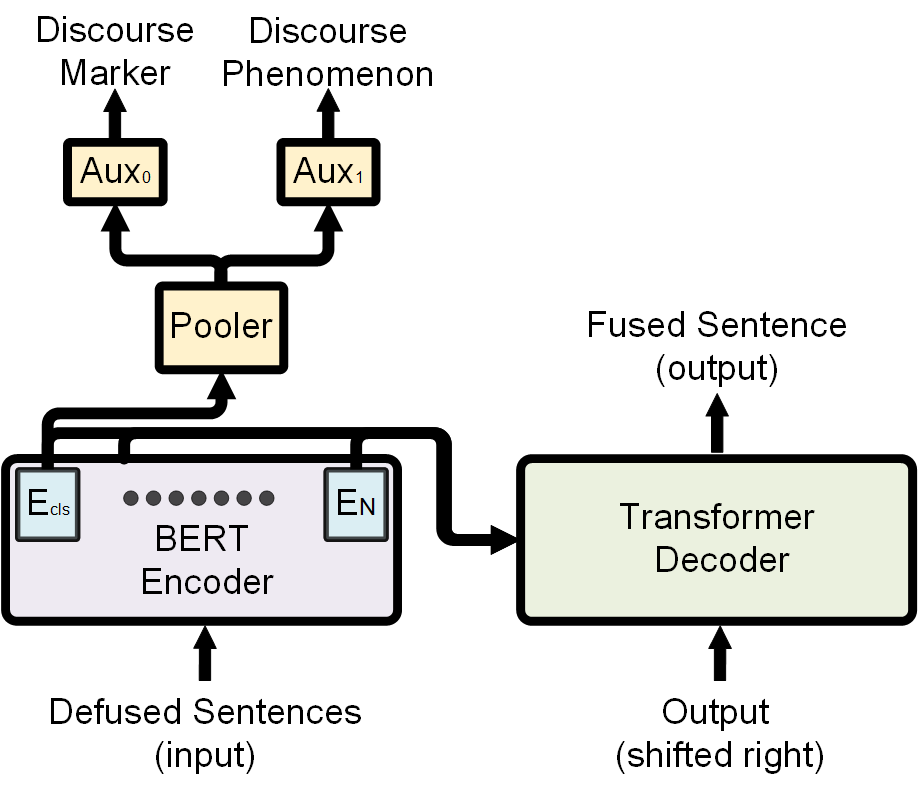}
\centering
\label{fig:classical_BERT}
\caption{The \AuxBertTrans{} architecture. $Aux_0$ and $Aux_1$ are classification layers for our auxiliary tasks.}
\centering
\label{fig:Aux-BERT-Transformer}
\end{figure}

The \AuxBertTrans{} architecture is depicted in \Cref{fig:Aux-BERT-Transformer}. Both the fusion task and the two auxiliary classification tasks share the contextualized representation provided by the BERT encoder. Each classification task has its own output head, while the fusion task is performed via a Transformer decoder.
The token-level outputs of the BERT encoder are processed by the attention mechanism of the Transformer decoder. BERT's CLS token, which provides a sentence-level representation, is post-processed by the \emph{pooler} (following \citealp{devlin2018bert}) and is fed to the two classification heads. The fusion component of the model 
%-- BERT encoding and Transformer decoding -- 
is identical to  \citet{rothe2019leveraging} (\BertTrans{}).

The three tasks we employ are combined in the following objective function: 
\begin{equation*} \label{eq:3}
\ell_{\total} = \ell_{\gen} + \alpha\cdot \ell_{\type} + \beta\cdot \ell_{\connec}
\end{equation*}
where $\ell_{\gen}$ is the cross-entropy loss of the generation task, while $\ell_{\type}$ and $\ell_{\connec}$ are the cross-entropy losses of the discourse type and connective phrase predictions, respectively, with scalar weights $\alpha$ and $\beta$. We utilize a pre-trained BERT encoder and fine-tune only its top two layers.

% We next describe how we automatically augment the examples in \disco{} with their semantically similar variants. As noted above, we do that in order to avoid arbitrary training signals, by explicitly encoding the semantic similarity of some of the discourse markers. 

\section{Experimental Setup}\label{sec:settings}
%We conduct an extensive experimental evaluation, comparing our models to the current state-of-the-art fusion models. 
We follow prior work and use the balanced version of \disco{}
%, which down-sampled examples containing `and', `but', or anaphora 
(\S \ref{sec:related_work}).
The dataset consists of 4.5M examples for Wikipedia (`W') and 12M examples for Sports (`S'), split to 98\% training, 1\% test and 1\% dev. % We mark the Wikipedia domain with `W' and the Sports domain with `S'. 
We evaluate fusion models both in in-domain and cross-domain settings (training in one domain and testing on the other domain). We denote with {W $\rightarrow$ S} the setup where training is done on Wikipedia and testing on Sports, and similarly use {S $\rightarrow$ W} for the other way around.

We evaluate the following fusion models:\footnote{Relevant code URLs are in the supplementary material.}
% \roi{Did we use any publicly available code for which we should report a URL (e.g.in the appendix, but stating here that the URL can be found there) ? One may ask that for \Trans{}, \BertTrans{} and \LaserTagger{}.}

\begin{description}
    \item[\Trans{}] - the Transformer-based model by \citet{geva2019discofuse}.
    \item[\LaserTagger{}] - the sequence tagging model by \citet{malmi2019encode}.
    \item[\BertTrans{}] - the BERT-based state-of-the-art model by \citet{rothe2019leveraging}.
    \item[\AugBertTrans{}] - \BertTrans{} trained on our augmented MR training set (\S \ref{sec:augmentation}).
    \item[\AuxBertTrans{}] - Our multitask model (\S \ref{sec:auxiliary model}).
    \item[\AugAuxBertTrans{}] - Our multitask model trained on our augmented MR training set  (\S \ref{sec:auxiliary model}).
\end{description}
%
%We note that all the above models whose name do not contain the `Aug' prefix are trained on the original non-augmented training set.

All baselines used the same parameter settings described in the cited works, and our models follow the parameter settings in \citet{rothe2019leveraging}. The same batch size and number of training steps were used in all models, thus training on the same number of examples when using either the original \disco{} or our augmented version.
The $\alpha$ and $\beta$ hyper-parameters of the multi-task objective are tuned on the dev set (see the supp. material).
% \roi{What do we mean by "the same training steps" ? Same number ? Other things ?} 

%\eric{What do we use for $\alpha$ and $\beta$ and did we optimize those on the development set?}\eyal{0.1, we did tune them on the validation set. Though we did not explore this intensively. I will elaborate on this in the appendix.}

% \begin{table}[t]
% \small
% \centering
% \begin{tabular}{ @{}C{1.95cm} | @{}C{1.5cm}@{} | @{}C{1.5cm}@{} | @{}C{1.5cm}@{} | @{}C{1.5cm}@{} }
% & {W } & {W $\rightarrow$ S} & {S} & {S $\rightarrow$ W}\\
%     \hline
%     \LaserTagger & 53.6 / 56.1 & 49.8 / 51.2 & 58.4 / 59.7 & 49.7 / 51.2\\
%      \BertTrans & 63.9 / 67.9 & 55.5 / 57.4 & 60.6 / 63.2 & 55.9 / 59.7 \\ 
%         \hline
%      \AuxBertTrans & \textbf{65.0} / 68.5 & \textbf{56.5} / 58.1 & \textbf{61.8} / 64.2 & \textbf{57.0} / 61.2 \\
%         % \hline
%      \AugBertTrans & 53.0 / 69.9 & 46.5 / 59.7 & 51.7 / 65.7 & 46.2 / 62.7 \\ 
%      %   \hline
%      \AugAuxBertTrans & 53.7 / \textbf{71.0} & 47.7 / \textbf{60.9} & 52.9 / \textbf{67.1} & 47.3 / \textbf{63.9}\\
% \end{tabular}
% \caption{Model performance with the \SoftEXACT{} metric.}
% \label{tab:Soft-EXACT 4 fold}
% \end{table}

\begin{table}[t!]
\small
\centering
\begin{tabular}{ c | c | c | c | c }
& {W } & {W $\rightarrow$ S} & {S} & {S $\rightarrow$ W}\\
    \hline
    \LaserTagger & 56.1 & 51.2 & 59.7 &  51.2\\
     \BertTrans & 67.9 & 57.4 & 63.2 & 59.7 \\ 
        \hline
     \AugBertTrans & 69.9 & 59.7 & 65.7 & 62.7 \\ 
     %   \hline
     \AuxBertTrans & 68.5 & 58.1 & 64.2 & 61.2 \\
        % \hline
     \AugAuxBertTrans & \textbf{71.0} & \textbf{60.9} & \textbf{67.1} &  \textbf{63.9}\\
\end{tabular}
\caption{Multi reference \EXACT{} (\MREXACT) results.}
\label{tab:MRP-EXACT 4 fold}
\end{table}

\begin{table}[t!]
\small
\centering
\begin{tabular}{ c | c | c | c | c }
& {W} & {W $\rightarrow$ S} & {S} & {S $\rightarrow$ W} \\
    \hline
    \LaserTagger & 79.8 & 79.5 & 82.7 & 77.6 \\
     %   \hline
     \BertTrans  & 90.3 & 86.4 & 88.0 & 86.6 \\ 
     \hline
     \AugBertTrans & 90.7 & 86.5 & 88.7 & 86.8 \\
     \AuxBertTrans & 90.6 & \textbf{87.0} & 88.4 & 86.7 \\
     \AugAuxBertTrans & \textbf{91.1} & \textbf{87.0} & \textbf{89.2} & \textbf{87.0} \\

\end{tabular}
\caption{Multi reference \SARI{} (\MRSARI) results.}
\label{tab:MRP-SARI 4 fold}
\end{table}

\begin{table}[t!]
\small
\centering
\begin{tabular}{ c | c | c | c | c }
& {W } & {W $\rightarrow$ S} & {S} & {S $\rightarrow$ W} \\
    \hline
     \Trans & 51.1 & 40.1 & 50.6 & 41.9 \\
     \LaserTagger & 54.6 & 49.8 & 58.4 & 49.7 \\
     %   \hline
     \BertTrans & 63.9 & 55.5 & 60.6 & 55.9 \\ 
        \hline
     \AugBertTrans & 53.0 & 46.5 & 51.7 & 46.2 \\
     \AuxBertTrans & \textbf{65.0} & \textbf{56.5} & \textbf{61.8} & \textbf{57.0} \\
     \AugAuxBertTrans & 53.7 & 47.7 & 52.9 & 47.3 \\
\end{tabular}
\caption{Single reference \EXACT{} results.}
\label{tab:EXACT 4 fold}
\end{table}

\begin{table}[t!]
\small
\centering
\begin{tabular}{ c | c | c | c | c }
& {W} & {W $\rightarrow$ S} & {S} & {S $\rightarrow$ W} \\
    \hline
    \Trans & 84.5 & 80.1 & 83.9 & 80.0\\
    \LaserTagger & 79.1 & 78.6 & 81.9 & 76.4\\
     %   \hline
     \BertTrans  & 89.2 & 85.8 & 87.2 & 85.3\\  % & 89.2 & 82.4 & 87.2 & 83.0
     \hline
     \AugBertTrans & 85.3 & 81.9 & 83.9 & 82.5 \\  % & 85.3 & 81.9 & 83.9 & 82.5 \\
     \AuxBertTrans& \textbf{89.5} & \textbf{86.0} & \textbf{87.6} & \textbf{85.5} \\ % \ & 89.5 & 82.8 & 87.6 & 83.4
     \AugAuxBertTrans & 85.7 & 82.5 & 84.4 & 82.9 \\ % & 85.7 & 82.5 & 84.4 & 82.9
 
\end{tabular}
\caption{Single reference \SARI{} results.}
\label{tab:SARI 4 fold}
\end{table}

\begin{table}[t]
\small
\centering
% \begin{tabular}{@{}p{3.15cm}|c|c||c|c}
\begin{tabular}{@{}p{3.3cm} |c|c||c|@{}c@{}}
    & \multicolumn{2}{c||}{\textbf{\BertTrans}} & \multicolumn{2}{c@{} }{\textbf{\AugAuxBertTrans{}}} \\
    \textbf{Discourse phenomena} & S & W & S & W \\
    \hline
    VP coordination         & 68.5 & 67.1 & 67.9 & 67.8 \\
    %   \hline
    Inner connective                 & 66.0 & 71.3 & \textbf{69.3} & \textbf{74.4} \\
    %   \hline
    Inner connective+A      & 46.0 & 58.0 & \textbf{52.6} & \textbf{61.0} \\ 
    %   \hline
    Sentence coordination            & 52.1 & 56.4 & \textbf{59.4} & \textbf{63.5} \\
    %   \hline
    Sentence coordination+ A & 32.0 & 40.1 & \textbf{42.1} & \textbf{48.6} \\
    %   \hline
    Forward connective               & 61.7 & 82.8 & \textbf{67.2} & 81.9 \\
    %   \hline
    Discourse connective             & 29.6 & 49.0 & \textbf{48.0} & \textbf{61.6} \\
    %   \hline
    Discourse connective+A & 5.3 & 18.5 & \textbf{22.7} & \textbf{30.6} \\
      \hline
      \hline
      Total Semantic & 52.7 & 59.5 & \textbf{59.7} & \textbf{64.9} \\
      \hline
      \hline
    Apposition & 98.4 & 98.0 & 98.6 & 98.6 \\
    %  \hline
    Relative Clause & 90.9 & 91.1 & 91.9 & 89.4 \\
    %  \hline
    Cataphora & 91.5 & 94.0 & 90.6 & 94.0 \\
    %  hline
    None & \textbf{66.9} & 68.1 & 57.6 & 68.5 \\
    % hline
    Anaphora & \textbf{62.0} & 62.5 & 58.4 & 61.9 \\
    \hline
    \hline
    Total Entity-centric & \textbf{82.5} & 83.0 & 80.6 & 82.8 \\
\end{tabular}
\caption{In-domain evaluation with \MREXACT, split by fusion phenomena. Boldfaced figures represent big gaps (more than 1.5\%) between models. '+A' indicates an addition of the anaphora phenomenon.}
% \eric{Please specify a threshold for a gap above which you use boldface or simply use boldface on every row.} 

\label{tab:sliced}
\end{table}

\begin{table}[t]
\centering
\small
\begin{tabular}{@{}p{1.9cm} | @{ }c@{ } | @{ }c@{ } | @{ }c@{ } | @{ }c@{}}
\multicolumn{5}{c }{\textbf{Sports}}\\
\hline
% & \textbf{Conjunction} & \textbf{Comparison} & \textbf{Cause} & \textbf{Relative} \\
& Conjunction & Comparison & Cause & Relative \\
    \hline
     \BertTrans       & 47.0 & 52.3 & 33.0 & 90.9\\ 
    % \hline
    \AugBertTrans   & 46.4 & 74.5 & 41.6 & 90.7 \\
    % \hline
     \AuxBertTrans   & 47.4 & 53.5 & 33.7 & \textbf{91.9} \\
    % \hline
     \AugAuxBertTrans & \textbf{47.7} & \textbf{74.7} & \textbf{43.6} & \textbf{91.9}\\
\end{tabular}
\newline
\newline
\newline
\begin{tabular}{@{}p{1.9cm} | @{ }c@{ } | @{ }c@{ } | @{ }c@{ } | @{ }c@{}}
\multicolumn{5}{c }{\textbf{Wikipedia}}\\
\hline
% & \textbf{Conjunction} & \textbf{Comparison} & \textbf{Cause} & \textbf{Relative}\\
& Conjunction & Comparison & Cause & Relative \\
    \hline
     \BertTrans       & 55.0 & 67.1 & 33.3 & \textbf{91.1}\\ 
    % \hline
    \AugBertTrans   & 55.2 & 75.1 & 43.6 & 89.6 \\
    % \hline
     \AuxBertTrans   & \textbf{56.7} & 67.7 & 39.6 & 90.9\\
    % \hline
     \AugAuxBertTrans & 56.6 & \textbf{76.0} & \textbf{46.0} & 89.4\\
\end{tabular}
\caption{Model performance across semantic clusters, measured with \MREXACT.}
\label{tab:Soft-EXACT semantic}
\end{table}

\begin{table*}[t]
\centering
%% local settings
\small
% for avoiding siunitx using bold extended
\renewrobustcmd{\bfseries}{\fontseries{b}\selectfont}
\renewrobustcmd{\boldmath}{}
% abbreviation
\newrobustcmd{\B}{\bfseries}

\begin{tabular}{c l c c }
 \hline
& \textbf{Fusion} & \textbf{BERTScore} & \textbf{\MREXACT{}} \\
    \hline
    % \\
     \textcolor{violet}{(R)} & Ruby is the traditional birthstone for July and is usually more pink than garnet\textbf{,}  & - & - \\
     & \textbf{ however} some rhodolite garnets have a similar pinkish hue to most rubies . & & \\
    %  \\
     \textcolor{darkgreen}{(G)} & \textbf{Although} ruby is the traditional birthstone for July and is usually more pink  , &  0.9670 & \textbf{1} \\
    & than garnet some rhodolite garnets have a similar pinkish hue to most rubies. &  & \\
    % \\
    \textcolor{red}{(B)} & Ruby is the traditional birthstone for July and is usually more pink than garnet \textbf{,}  & \textbf{0.9893} & 0 \\
     & \textbf{thus} some rhodolite garnets have a similar pinkish hue to most rubies . & & \\
    % \\ 
    \hline
    % \\
     \textcolor{violet}{(R)} & The water level in the wells has risen. \textbf{As a result,} work on agricultural lands & - & - \\
     & is going on. & & \\
    %  \\
     \textcolor{darkgreen}{(G)} & The water level in the wells has risen\textbf{, hence} work on agricultural lands &  0.9713 & \textbf{1} \\
     & is going on. & & \\
    % \\
    \textcolor{red}{(B)} & The water level in the wells has risen. \textbf{However,} work on agricultural lands & \textbf{0.9745} & 0 \\
    & is going on. & & \\
    
    % \\
    \hline
    % \\
     \textcolor{violet}{(R)} & August 28, \textbf{which is the} second day after school starts, is their first away game. & - & - \\
    %  \\
     \textcolor{darkgreen}{(G)} & august 28, \textbf{the} second day after school starts, is their first away game. &  0.9834 & \textbf{1} \\
    
    % \\
    \textcolor{red}{(B)} & August 28, \textbf{who is the} second day after school starts, is their first away game. & \textbf{0.9879} & 0 \\
    %  \\
    \hline
\end{tabular}
\caption{A demonstration of BERTScore \citep{bert-score} and \MREXACT{} (ours) evaluation scores for sentence fusion examples. We mark the ground-truth reference fusion with \textcolor{violet}{(R)}, a correct variant with \textcolor{darkgreen}{(G)} and an incorrect variant with \textcolor{darkred}{(B)}.}
\label{tab:BERTScore-mistakes}
\end{table*}

% \begin{table*}
% \begin{center}
% \small
% \begin{tabular}{ l c c}
%  \hline
% \textbf{Fusion} & \textbf{\BertTrans} & \textbf{\AugBertTrans} \\
%     \hline
%      He began his audio experimentations. \textbf{Consequently,} he flunked his first semester of college. & 0.0001 & 0.0017 \\
%      He began his audio experimentations. \textbf{As a result,} he flunked his first semester of college. & 0.0047 & 0.0024 \\
%      He began his audio experimentations. \textbf{Hence,} he flunked his first semester of college. & 0.0000 & 0.0017 \\
%      He began his audio experimentations. \textbf{Thus,} he flunked his first semester of college. & 0.0001 & 0.0019 \\
%      He began his audio experimentations. \textbf{Therefore,} he flunked his first semester of college. & 0.0000 & 0.0022 \\
%     \hline
%     \hline

% \end{tabular}
% \caption{Example of fusion variants distribution according to a baseline model and a model that is trained on our augmented \disco{}, \BertTrans{} and \AugBertTrans{} respectively. The values in right columns represent the probability given to a sentences according to the model. The first sentence is the original fusion, while the others are variants we have generated. }
% \label{tab:sentence-probs}
% \end{center}
% \end{table*}

% \begin{figure*}
% \includegraphics[scale=0.28]{figures/sentence-probs.png}
% \centering
% \caption{Fusion variants distribution according to \BertTrans{} and \AugBertTrans{}. E stands for the distribution entropy and P stands for the sum of variants probabilities.}
% \centering
% \label{fig:sentence-probs}
% \end{figure*}

\section{Results}
\label{sec:results}

We report results with \MREXACT{} (\Cref{tab:MRP-EXACT 4 fold}) and \MRSARI{}  (\Cref{tab:MRP-SARI 4 fold}). To maintain compatibility with prior work, we also report results with single reference (SR) \EXACT{} (\Cref{tab:EXACT 4 fold}) and \SARI{}  (\Cref{tab:SARI 4 fold}).
Boldfaced figures in the tables are statistically significant with $p<0.001$ compared to the second best model (using McNemar's paired test for \EXACT{} and the Wilcoxon signed-rank test for \SARI{} \citep{dror2018hitchhiker}). 

For the SR evaluation, our \AuxBertTrans{} is best performing, indicating the value of our multitask framework. On the other hand, training with the augmented dataset has a negative impact. This is not surprising since SR evaluation uses one arbitrary reference, while the augmented dataset guides the model towards balanced fusion variants.
%
%We argue that the negative impact of augmented training under this evaluation stems from the following reason: With the semantically balanced training sets the trained models offer more diverse fusion variants, a property that we consider highly desirable, but that do not coincide with the distribution of semantic relations in the original test sets. 
%
%
Our premise in this paper is that multi-reference evaluation is more adequate in assessing outcomes that paraphrase the original \disco{} fusions.
%
%As discussed in \Cref{sec:multi_ref}, MRP based evaluation is more adequate in measuring fusion outcomes that could be paraphrases of the gold fusions.
Indeed, the results in Tables \ref{tab:MRP-EXACT 4 fold} and \ref{tab:MRP-SARI 4 fold} show that with MR evaluation all our models outperform all baselines across setups, with \AugAuxBertTrans{}, which combines multi-reference training and semantic guidance using auxiliary tasks, performing best. 

We further analyze in \Cref{tab:sliced} the in-domain model performance of the strongest baseline \BertTrans{} and our strongest model \AugAuxBertTrans{} using \MREXACT, sliced by the different discourse phenomena annotated in \disco{}. 
As discussed in \S \ref{sec:bert-lim}, we distinguish between two fusion phenomena types. \emph{Entity-centric} fusion phenomena bridge between two mentions of the same entity, and for such, no connective discourse marker should be added by the model.
%As discussed in \S \ref{sec:bert-lim}, we distinguish between what we perceive as \emph{entity-centric} fusion phenomena and as \emph{semantic relationship} fusion phenomena. The former cases bridge between two mentions of the same entity in the two input sentences, and as such, no connective discourse marker should be generated by the model.
%and as such they do not revolve around the semantic relation between the two input sentences but around several information pieces for the same entity. They tend to be entity-centric in the sense that they bridge between two mentions of the same entity in the two input sentence.
Our analysis shows that both models perform well on 3 of the 5 \emph{entity-centric} phenomena (bottom part of \Cref{tab:sliced}). For None and Anaphora, there is a drop in \AugAuxBertTrans{} performance, which may be attributed to the change in example distribution introduced by our augmented dataset, and will be addressed in future work.
%Examples are the bottom 5 phenomena (apposition, relative clause, cataphora, anaphora, and `none').

The \emph{semantic relationship} phenomena, on the other hand, require deeper understanding of the relationship between the two input sentences. They tend to be more difficult as they involve the choice of the right connective according to this relation.
On these phenomena (top part of \Cref{tab:sliced}), \AugAuxBertTrans{} provides substantial improvements compared to \BertTrans{}, indicating the effectiveness of guiding a model towards a robust semantic interpretation of the fusion task via multiple references and multitasking. Specifically, in the most difficult phenomenon for \BertTrans{}, discourse connectives, performance increased relatively by 62\% for Sports and 26\% for Wikipedia. The gap is even larger for the composite cases of discourse connectives combined with anaphora (``Discourse connective+A''): 328.3\% (Sports) and 65.4\% (Wikipedia). 

Finally, to explore the relative importance of the different components of our model, we looked at model performance sliced by the clusters we introduced in \S \ref{ssec:fusion_variants} (see \Cref{tab:Clustering-connectives}). The results (\Cref{tab:Soft-EXACT semantic}), show that 
\AuxBertTrans{} outperforms \BertTrans{} in 7 of 8 cases, but the gap is $\leq 2$\% in all cases but one. On the other hand, \AugBertTrans{} improves over \BertTrans{} mostly for \qex{Comparison} and \qex{Cause}, but the average improvements on these clusters are large: 15.4\% (Sports) and 9.2\% (Wikipedia). This shows that while our auxiliary tasks offer a consistent performance boost, the inclusion of multiple references contribute to significant changes in model's semantic perception.

%\AugAuxBertTrans{} is the best performing model in 6 of 8 cases, and that its gap from baseline \BertTrans{} model is $ \geq 8.9$\% for the \qex{Comparison} and \qex{Cause} clusters. \AugBertTrans{} also outperforms \BertTrans{} by large margins on these clusters, while \AuxBertTrans{} shows a $\geq 1.5$\% improvement only for the \qex{Cause} cluster in Sports. This emphasizes the importance of augmented training and the \qex{Comparison} and \qex{Cause} clusters for our results.

\section{Ablation Analysis}
\label{sec:qual}
In this analysis, we focus on potential alternative evaluation measures. As mentioned in \S \ref{sec:related_work}, a possible direction for solving issues in evaluation of sentence fusion---stemming from having a single reference---could be to use similarity-based evaluation metrics \citep{sellam2020bleurt,DBLP:conf/icml/KusnerSKW15, clark-etal-2019-sentence,bert-score}. We notice two  limitations in applying such metrics for sentence fusion. First, similarity-based metrics depend on trainable models that are often in use within fusion models. Thus, we expect these metrics to struggle in evaluation when fusion models struggle in prediction. Second, these metrics fail to correctly account for the semantics of discourse markers, which is particularly important for sentence fusion.

In \Cref{tab:BERTScore-mistakes} we illustrate typical evaluation mistakes made by BERTScore \citep{bert-score}, a recent similarity-based evaluation measure. We calculate BERTScore (F1) for each reference fusion with two variants; (1) a fusion that holds the same meaning and (2) a fusion with a different meaning. A valid evaluation measure for the task is supposed to favor the first option (i.e. the fusion with the same meaning). However, that is not the case for the given examples. The measure often fails to consider the semantic differences between sentences, which is an important element of the task.  

Consider the second example in \Cref{tab:BERTScore-mistakes}: BERTScore favours the structural similarity between the gold reference (\textcolor{violet}{R}) and the incorrect variant (\textcolor{darkred}{B}), which differ in meaning (yet based around the same fusion phenomenon: discourse connective). Meanwhile, the correct variant (\textcolor{darkgreen}{G})
holds the same meaning as the reference (while  a different fusion phenomenon is being used: sentence coordination instead of discourse connective).  
% The reference fusion (\textcolor{violet}{R}) 
% % states that the rise of the water level in the wells contributed to the on-going work on agricultural land. 
% and the correct variant (\textcolor{darkgreen}{G}) hold the same meaning, yet a different fusion phenomenon is being used (sentence coordination instead of discourse connective, see \Cref{tab:sliced}). 
% In contrast, the incorrect variant (\textcolor{darkred}{B}) 
% % depicts the rise in water levels in wells as an impediment to agricultural work, 
% has a different meaning while using the same fusion phenomena as in the gold-reference.
% % (discourse connective). 
% % A valid evaluation measure should prefer the correct variant and give it a higher score than that of the incorrect variant.  
% However, BERTScore favours the structural similarity between the gold reference and the incorrect variant over the semantic similarity between the gold reference and the correct variant.

\section{Conclusions}

We studied the task of sentence fusion and argued that a major limitation of common training and evaluation methods is their reliance on a single reference ground-truth. To address this, we presented a method that automatically expands ground-truth fusions into multiple references via curated equivalence classes of connective terms. We applied our method to a leading resource for the task.
%, expanding it to consider multiple references for each input example.
%commonly used evaluation metrics is failing to identify valid paraphrases of the ground-truth fusions. To address this limitation, we introduced the \SoftEXACT{} metric, which offers better semantic analysis of fusion results by automatically clustering semantically equivalent fusion variants.
%We showed that the state-of-the-art fusion model lacks in modeling the semantic relationship between the input sentences to be fused. 

We then introduced a model that utilizes multiple references by training on each reference as a separate example while learning a common semantic representation for surface form variances using auxiliary tasks for semantic relationship prediction in a multitasking framework.
Our model achieves state-of-the-art performance across a variety of evaluation scenarios.

Our approach for evaluating and training with generated multiple references is complementary to an approach that uses a similarity measure to match between similar texts. In future work, we plan to study the combination of the two approaches.

%We define a scheme to automatically augment more valid fusions to each example in discofuse. We use this scheme and augment more data, then we utilize this data to train our model. With the augmented fusions, we define a new evaluation metric for the task, which we call Soft-EXACT. We show that the our model that was trained on the augmented data, achieves best performance with Soft-EXACT evaluation.

%We provide intuition to the superiority of the Soft-EXACT evaluation metric on this task. Following an extensive analysis of model errors, we support this intuition with empirical results. This analysis also shed more light on the true nature of our model, demonstrating that it actually perform better than is revealed when using EXACT and SARI, which are common sentence fusion evaluation metrics. 

%While our model improved the modeling of fusion semantics, it still struggles to handle examples with a combination of several fusion phenomena. In future work, we plan to study how to guide a fusion model to handle these complex cases. In addition, our augmented training set changes the data distribution by balancing between similar fusion variants. We plan to investigate if this approach can be effective in other NLP tasks.

\section*{Acknowledgments}
We  would like to thank the members of the 
 IE@Technion NLP group and Roee Aharoni, for their valuable feedback and advice. 
 Roi Reichart was partially funded by ISF personal grant No. 1625/18.

% \bibliography{anthology, emnlp2020}
\bibliography{main}
\bibliographystyle{acl_natbib}

\appendix
% \section{Appendices}
% \label{sec:appendix}
%\section{Supplemental Material}
%\label{sec:supplemental}

\section{Augmentation Rules}
\label{sec:apend-augmentation-rules}

In this section we provide the technical details of the augmentation rules used to augment \disco{} (see \S \ref{ssec:fusion_variants}). For the sake of clarity, we only provide a general explanation of most rules, avoiding fine-grained issues, minor implementation details and repeating similar rules with minor differences. We note that our augmented dataset will be made publicly available upon acceptance of the paper.

Given a triplet ($s_i^1$, $s_i^2$, $t_i$), where $s_i^1$ and $s_i^2$ are the input sentences and $t_i$ is the ground truth fusion, and its corresponding discourse phenomenon and marker, $p_i$ and $c_i$, respectively, we consider the semantic relationship in $t_i$ which is expressed by $c_i$ (see beginning of \cref{tab:rule_example}). Our augmentation rules relate to three semantic classes: Conjunction, Comparison and  Cause, and to one syntactic class: Relative clause (see class definition, \S \ref{ssec:fusion_variants}). We design a set of rules for each of these classes, such that each rule first specifies how to detect a fusion that can be augmented according to the rule, and then describes which operations to perform on the ground-truth fusion and its inputs, $t_i$, $s_i^1$ and $s_i^2$, in order to generate a new valid fusion. 

\begin{table}[t]
\small
\centering
\begin{tabular}{ l  l }
\textbf{Dict} & \textbf{Key}  \\
    \hline
    $\mathcal{C}_{a}$ & ``furthermore'' , ``moreover'' , ``additionally'' , \\ 
    \hline
    $\mathcal{C}_{,a}$ & ``, and'' \\
    \hline
    $\mathcal{C}_{a,}$ & ``furthermore,'' , ``plus,'' , ``additionally,'' , \\ 
          &  ``moreover,''   \\
    \hline
    $\mathcal{C}_{q}$ &  ``however'' , ``yet'' , ``but'', ``nevertheless'' ,\\                  & ``although'' \\
    \hline
    $\mathcal{C}_{,q}$ &  ``, yet'' , ``, but'', ``although'' \\       
    \hline
    $\mathcal{C}_{q,}$ &  ``however,'' , ``still,'', ``although,'' \\        
    & ``nevertheless,''  \\
    \hline
    $\mathcal{C}_{e}$ &  ``hence'' , ``therefore'' , ``consequently'' \\
    \hline
    $\mathcal{C}_{e,}$ &  ``as a result,'' , ``hence,'' , ``thus,'' ,  \\                   & ``consequently,'' , ``therefore,'' \\
    \hline
    $\mathcal{E}_r$ & ``who is a'', ``who is not a'', ``who is an'', \\ 
                    & ``who are an'', ``who are a'', ``who is the'', \\
                    & ``who is not an'', ``who are not a'', \\
                    & ``who are the'', ``which is a'', ``which is an'', \\
                    & ``who are not an'', ``who is not the'', \\
                    & ``who are not the'', ``which is not a'', \\
                    & ``which are an'', ``which are a'', ``which is the''\\
                    & ``which is not an'', ``which are not a'', \\
                    & ``which are not an'', ``which are the'', \\
                    & ``which is not the'', ``which are not the'' \\
    \hline
    $\mathcal{P}_r$ & ``who is'', ``who are'', ``which is'', ``which are'' \\ 
    \hline
    \end{tabular}
\caption{The dictionaries used in the data augmentation process.}
\label{tab:augmentation disctionries}
\end{table}

\begin{table}[ht!]
\small
\centering
\begin{tabular}{l | l}
\textbf{Operation} & \textbf{Description} \\
\hline
Replace($T$, $s_0$, $s_1$) & Replace occurrences of $s_0$ in $T$\\
& with $s_1$.\\
Concat($s_0$, $s_1$) & Attach the string $s_1$ to the end \\
& of $s_0$. \\
Delete($T$, $s_0$) &  Delete occurrences of $s_0$ from\\
& $T$.\\
\hline
\end{tabular}
\caption{Operations on sentences and text phrases, applied for data augmentation (the actual rules are in \Cref{tab:detailed augmentation rules}). $T$, $s_0$ and $s_1$ are strings, where $T$ is often an entire sentence while $s_0$ and $s_1$ are phrases.}
\label{tab:operations}
\end{table}

\begin{table*}[t]
\small
\centering
\begin{tabular}{ l | l | l@{}}
\toprule
\textbf{Semantic} & \textbf{Detection} & \textbf{Augmentation} \\
    \hline
    Conjunction & $c_i, c' \in\hat{\mathcal{C}_{a}}$ & Replace($t_i$, $c_i$, $c'$) \\
             & $c_i\in\mathcal{C}_{a} \wedge  c'\in\mathcal{C}_{a,}\wedge \text{Concat}(.,c_i)\in{t_i}$ & Replace($t_i$, $c_i$, $c'$) \\
             & $c_i\in\mathcal{C}_{a} \wedge  c'\in\mathcal{C}_{,a}\wedge \text{Concat}(., c_i)\in{t_i} \wedge len(t)<40\ $ & Replace($t_i$, Concat(``.'', $c_i$), $c'$) \\
    \hline
    Comparison & $c_i, c'\in\hat{\mathcal{C}_{q}}  $ & Replace($t_i$, $c_i$, $c'$) \\
    & $ c_i \not\in\mathcal{C}_{q,}\wedge c'\in \mathcal{C}_{q,}\wedge p=\text{S-coordination}\;  $ & Replace($t_i$, $c_i$, Concat(``.'', $c'$)) \\
    & $c_i\not\in{\mathcal{C}_{,q}} \wedge c' \in{\mathcal{C}_{,q}} \wedge p=\text{Inner-connective}\; \wedge c'\in \{\text{but, yet}\} $ & Replace($t_i$, $c_i$, $c'$)\\
    & $c_i\not\in{\mathcal{C}_{q,}} \wedge c' \in{\mathcal{C}_{q,}} \wedge p=\text{Inner-connective} $ & Replace($t_i$, $c_i$, Concat(``.'', $c'$)) \\
    & $c_i\not\in{\mathcal{C}_{q,}} \wedge c' \in{\mathcal{C}_{q,}} \wedge p=\text{Discourse-connective} $ & Replace($t_i$, $c_i$, $c'$) \\
    & $c_i\not\in{\mathcal{C}_{,q}} \wedge c'=\{\text{, although}\} \wedge p=\text{Discourse-connective} $ & Replace($t_i$, Concat(``.'',$c_i$), $c'$)\\
    & \textcolor{red}{$c'\in{\mathcal{C}_{,q}} \wedge c_i=\{\text{although}\} \wedge p=\text{Forward-connective} $} & \textcolor{red}{Concat(Concat(Delete($s_i^1$, ``.''), $c'$), $s_i^2$)}\\
    \hline
    Cause & $c_i, c' \in\hat{\mathcal{C}_{e}}$ & Replace($t_i$, $c_i$, $c'$)\\
           & $c_i \in \mathcal{C}_{e} \wedge c' \in \mathcal{C}_{e,} \wedge \text{Concat}(., )\in t_i \wedge \text{Concat}(.,c_i)\not\in t_i $ & Replace(Delete($t_i$, $c_i$), ``.'', Concat(``.'',$c'$))\\             
    \hline
    Relative & $p=\text{Relative Clause} \wedge \exists a\in \mathcal{E}_r\; |\; a\in t_i$  & Delete($t_i$, $b$), $b\in a\cap \mathcal{P}_r$ \\
    Clause &   &  \\
\toprule
\end{tabular}
\caption{Augmentation rules for derivations of new fusions out of \disco{} ground-truth fusions. We mark with \textcolor{red}{red} the rule discussed in the detailed augmentation example in \Cref{tab:rule_example}.}
\label{tab:detailed augmentation rules}
\end{table*}

\begin{table}[ht!]
\small
\centering
\begin{tabular}{@{}| l  l |}
   \multicolumn{1}{l}{\textbf{Notation}} & \multicolumn{1}{l}{\textbf{Definition}} \\
    % \textbf{Notation} & \textbf{Definition} \\ 
    \hline
    $t_i$  & The ground-truth fusion of the $i$-th example \\
    \hline
    $s_i^1$, $s_i^2$  & The two input sentences of the $i$-th \\
    & example\\
    \hline
    $c_i$  & The discourse marker used in $t_i$ \\
    \hline
    $p_i$  & The discourse phenomenon of $t_i$ \\
    \hline
    \hline
    $\mathcal{C}_{a}$  & A list of conjunction markers, without\\
                       & a comma \\
    \hline
    $\mathcal{C}_{a,}$  & A list of conjunction markers with  \\                                      & a right comma \\
    \hline
    $\mathcal{C}_{,a}$  & A list of conjunction markers with \\                                       & a left comma \\
    \hline
    $\mathcal{C}_{q}$  & A list of comparison markers, without\\                                     & a comma \\
    \hline
    $\mathcal{C}_{q,}$  & A list of comparison markers with\\                                         & a right comma \\
    \hline
    $\mathcal{C}_{,q}$  & A list of comparison markers with\\                                         & a left comma \\
    \hline
    $\mathcal{C}_{e}$  & A list of cause and effect markers, \\                                   & without a comma \\
    \hline
    $\mathcal{C}_{e,}$  & A list of cause and effect markers \\                                    & with a right comma \\
    \hline
    $\mathcal{E}_r$  & A set of relative clause expressions which  \\
                     & can transform to an apposition phrases  \\
                     & without adding any tokens  \\
    \hline
    $\mathcal{P}_r$  & A set of relative clause pronouns \\
    \hline
\end{tabular}
\caption{Notations and definitions for \Cref{tab:detailed augmentation rules}.}
\label{tab:notations}
\end{table}

We use a set of dictionaries, depicted in \Cref{tab:augmentation disctionries}, and a list of pre-defined operations, depicted in \Cref{tab:operations}. 
In \Cref{tab:detailed augmentation rules} we provide the technicalities of each rule, presenting its detection and augmentation schema, which is accompanied by the notations and definitions provided in \Cref{tab:notations}.

In \Cref{tab:rule_example} we provide a detailed example of the augmentation process. We start with a description of the input structure, which is detected as a fit for an augmentation rule. We then demonstrate how the variant generation is performed, in a step by step manner.

\begin{table}
\begin{center}
\small{
\begin{tabular}{p{6.7cm}p{0.1cm}}
\hline
\multicolumn{2}{|l|}{\textbf{1. Input: Ground-truth fusion}} \\
\multicolumn{2}{|l|}{} \\
\multicolumn{2}{|p{6.8cm}|}{$s_i^1 = $ \{The company had bigger facilities at Wembley in the west of the capital.\}}
\\
\multicolumn{2}{|p{6.8cm}|}{$s_i^2 = $ \{It was easier to attract stars and audiences to central London.\}}
\\
\multicolumn{2}{|p{6.8cm}|}{$t_i = $ \{\textcolor{red}{Although} the company had bigger facilities at Wembley in the west of the capital, it was easier to attract stars and audiences to central London.\}} \\
\multicolumn{2}{|p{6.8cm}|}{$c_i=\text{although}$}\\
\multicolumn{2}{|p{6.8cm}|}{$p_i=\text{Forward-connective}$}\\
\multicolumn{2}{|l|}{} \\
\hline
\hline
\multicolumn{2}{|p{6.8cm}|}{\textbf{2. Detection}} \\
\multicolumn{2}{|p{6.8cm}|}{} \\
\multicolumn{2}{|p{6.8cm}|}{$c'=\{\text{, but}\}\in{\mathcal{C}_{,q}} \wedge c_i=\text{\{although\}} \wedge p_i=\text{Forward-connective}$} \\
\multicolumn{2}{|p{6.8cm}|}{} \\
\hline 
\hline
\multicolumn{2}{|p{6.8cm}|}{\textbf{3. Operations}} \\
\multicolumn{2}{|p{6.8cm}|}{} \\
\multicolumn{1}{|p{2.3cm}|}{$\textsc{Delete}(s_i^1, ``.'') $} & \multicolumn{1}{p{4.5cm}|}{$X_1 = $ \{The company had bigger facilities at Wembley in the west of the capital\}} \\
\multicolumn{1}{|p{2.3cm}|}{} & \multicolumn{1}{p{4.5cm}|}{} \\
\multicolumn{1}{|p{2.3cm}|}{$\textsc{Concat}(X_1,c')$} & \multicolumn{1}{p{4.5cm}|}{$X_2 = $ \{The company had bigger facilities at Wembley in the west of the capital\textcolor{red}{, but}\}} \\
\multicolumn{1}{|p{2.3cm}|}{} & \multicolumn{1}{p{4.5cm}|}{} \\
\multicolumn{1}{|p{2.3cm}|}{$\textsc{Concat}(X_2,s_i^2)$} & \multicolumn{1}{p{4.5cm}|}{$X_3 = $ \{The company had bigger facilities at Wembley in the west of the capital\textcolor{red}{, but} it was easier to attract stars and audiences to central London.\}} \\
\multicolumn{2}{|p{6.8cm}|}{} \\
\hline
\hline 
\multicolumn{2}{|p{6.8cm}|}{\textbf{4. Output - augmented fusion}} \\
\multicolumn{2}{|p{6.8cm}|}{} \\
\multicolumn{2}{|p{6.8cm}|}{$t'_i$ = \{The company had bigger facilities at Wembley in the west of the capital\textcolor{red}{, but} it was easier to attract stars and audiences to central London.\}} \\
\multicolumn{2}{|p{6.8cm}|}{$c_i'=\text{, but}$}\\
\multicolumn{2}{|p{6.8cm}|}{$p_i'=\text{Sentence-coordination}$}\\
\multicolumn{2}{|p{6.8cm}|}{} \\
\hline
\end{tabular}}
\end{center}
\caption{A detailed augmentation rule execution example. We mark  discourse markers in \textcolor{red}{red}. The ground-truth fusion $t_i$ consists of the input together with the two source sentences, $s_i^1$ and $s_i^2$.}
\label{tab:rule_example}
\end{table}

\section{Augmentation Statistics}

We provide statistics for the entries in our augmented dataset.  
\Cref{tab:markers augmentation distribution} and \Cref{tab:phenomena augmentation distribution} show the distributions of the augmented discourse markers and phenomena, respectively. We note that we have generated a total of 6.5M and 14.7M new fusion examples out of the balanced \disco{} dataset in the Wikipedia and Sports domains, respectively. We then sampled 5M examples of each domain for training \AugBertTrans{} and \AugAuxBertTrans{}. These tables provide details about the imbalanced augmentation, where specific phenomena and markers are generated more often than others during the augmentation process.  %This stems from the imbalanced difficulties of manually handcrafting new fusions to different classes.

\begin{table}[t]
\small
\centering
\begin{tabular}{| l | l | l | l |}
\multicolumn{2}{c}{\textbf{Sports}} & \multicolumn{2}{c}{\textbf{Wikipedia}}\\
\hline
 & \% &  & \%  \\
\hline
although & 18.8 & still & 15.1  \\
\hline
yet & 16.7 & although & 24.4  \\
\hline
nevertheless & 16.0 & nevertheless & 15.9  \\
\hline
still & 14.0 & however & 10.6  \\
\hline
however & 13.1 & yet & 15.6  \\
\hline
but & 8.4 & but & 8.4  \\
\hline
consequently & 1.1 & hence & 1.3  \\
\hline
moreover & 1 & consequently & 1.2  \\
\hline
\toprule
\end{tabular}
\caption{The most common connectives augmented to the balanced \disco{} dataset. Percentages are calculated with respect to the entire set of new fusions in each domain.}
\label{tab:markers augmentation distribution}
\end{table}

\begin{table}[t]
\small
\centering
\begin{tabular}{@{}|l|l|l|}
    \hline
    \textbf{Discourse phenomena} & \textbf{Sports (\%)} & \textbf{Wiki(\%)}\\
    \hline
    VP coordination         & 7.8 & 6.8 \\
      \hline
    Inner connective        & 9.5 & 6.1 \\
      \hline
    Inner connective+A      & 2.4 & 2.7 \\
      \hline
    Sentence coordination   & 12.1 & 11.2 \\
      \hline
    Sentence coordination+A & 3.1 & 4.3 \\
      \hline
    Discourse connective   & 48.9 & 44.8 \\
      \hline
    Discourse connective+A & 15.7 & 23.7 \\
      \hline
    Apposition & 0.2 & 0.1 \\
    \hline
    \toprule
\end{tabular}
\caption{Discourse phenomena distribution of augmented fusions from the balanced \disco{} dataset. '+A' indicates an addition (composition) of the anaphora phenomenon, and Wiki stands for Wikipedia.}
\label{tab:phenomena augmentation distribution}
\end{table}

\begin{figure*}[t!]
\includegraphics[scale=0.30]{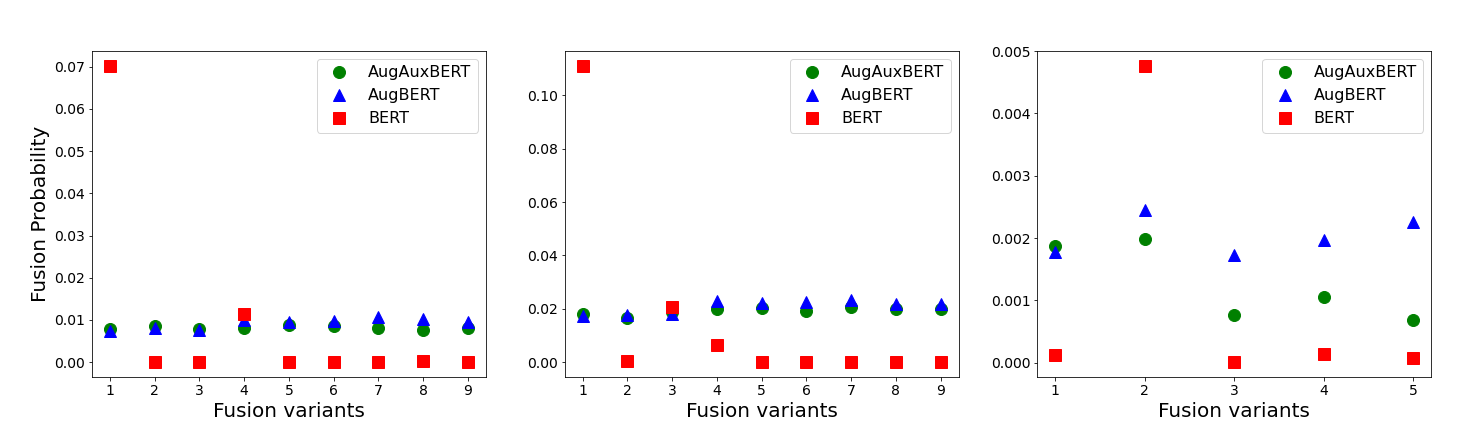}
\centering
\caption{The probability assigned to the ground-truth fusions by our \AugBertTrans{} and \AugAuxBertTrans{} models, and by the baseline \BertTrans{}, for three typical \disco{} examples. Our models assign higher and more uniform probabilities to the members of the ground-truth set.}
\centering
\label{fig:sentence-probs}
\end{figure*}

\begin{table}
\small
\centering
\begin{tabular}{l | l}
\toprule
\multicolumn{2}{c}{Encoder - BERT} \\
\hline
hidden size & 768 \\
number of attention heads & 12 \\
number of hidden layers & 12 \\
vocab size & 30522 \\
hidden activation & \emph{'gelu'} \\
number of parameters & 108891648 \\
\toprule
\multicolumn{2}{c}{Decoder - Transformer} \\
\hline
hidden size & 768 \\
number of attention heads & 8 \\
number of hidden layers & 6 \\
number of parameters & 47238144 \\
% beam width & 5 \\
\toprule
\multicolumn{2}{c}{Classifiers - Pooler} \\
\hline
input dim & 768 \\
first hidden dim & 768 \\
second hidden dim & 256 \\
phenomena output dim & 13 \\
connective output dim & 71 \\
number of parameters & 809044 \\
\toprule
\multicolumn{2}{c}{Optimization} \\
\hline
optimizer & \emph{AdamOptimizer} \\
beta1 & $0.9$ \\
beta2 & $0.997$ \\
epsilon & $1e-9$\\
batch size & 100 \\
\toprule
\end{tabular}
\caption{The hyper-parameters of the \BertTrans, \AuxBertTrans, \AugBertTrans{} and \AugAuxBertTrans{} models.}
\label{tab:models params}
\end{table}

\begin{table}[t]
\small
\centering
\begin{tabular}{ c | c | c }
& {W } & {S}\\
    \hline
     \BertTrans & 63.5 & 60.4 \\ 
        \hline
     \AuxBertTrans & \textbf{64.3} & \textbf{61.4} \\
        % \hline
     \AugBertTrans & 52.2 & 51.0 \\ 
     %   \hline
     \AugAuxBertTrans & 52.8 & 52.3 \\
\end{tabular}
\caption{Single reference EXACT results on development data.}
\label{tab:dev-results}
\end{table}

\section{Probability Distribution across Valid Fusions}
According to the results our models achieve in MR evaluation, we conclude that they are better capable of generating a fused text that is included in the ground-truth set. Here we show that, in addition, they assign a more uniform  probability to the members of the set, compared to the \BertTrans{} model. \Cref{fig:sentence-probs} graphically illustrates this pattern for three typical examples with 9, 9 and 5 ground-truth fusions, respectively (in each example, fusion 1 is the one in the original \disco{}, and the others were created in our expansion).

We first formally show that the probability mass tends to be uniformly allocated among the various references; for any  $t \in \mathcal{V}(t_i)$ let 
\begin{equation*}
\bar{p}_i(t) = \frac{p(t | s^1_i,s^2_i)}{\sum_{t' \in \mathcal{V}(t_i)}p(t' | s^1_i,s^2_i)}
\end{equation*}
be the probability of a variant $t$ relative to the overall probabilities of the variants in $\mathcal{V}(t)$. 
Indeed, for more than 99\% of the test-set examples the entropy $-\sum_{t \in \mathcal{V}(t_i)}\bar{p}_i(t)\log \bar{p}_i(t)\ $ induced by \AugBertTrans{} and \AugAuxBertTrans{} for the ground-truth solutions is higher than that of \BertTrans{}, indicating that our augmented models are less inclined to prefer one of the solutions over the others. 

% Indeed, for more than 99\% of the test-set examples the entropy of the probability distribution induced by \AugBertTrans{} and \AugAuxBertTrans{} for the ground-truth solutions is higher than that of \BertTrans{}, indicating that our augmented models are less inclined to prefer one of the solutions over the others. 

Moreover, we computed the sum of the multiple-reference probabilities $\sum_{t \in \mathcal{V}(t_i)}p(t | s^1_i,s^2_i)$ in test-set examples.
In about 71\% of the test-set examples the sum of  probabilities  induced by \AugBertTrans{} and \AugAuxBertTrans{} is higher than that of \BertTrans{}. 
That is, our model learns to direct more overall probability mass towards the correct variants, indicating a higher confidence in the correct solutions.

\section{Hyper-Parameters and Configurations}

The \BertTrans, \AuxBertTrans, \AugBertTrans{} and \AugAuxBertTrans{} models share the same hyper-parameters with respect to their shared architecture and to the optimization process. All models use an initialized BERT-Base Uncased  encoder  with a randomly initialized Transformer \citep{vaswani2017attention} decoder. Configuration details and the hyper-parameters of the training process are provided in \Cref{tab:models params}. 

Recall that we define our multi-task loss as follows:

\begin{equation*} \label{eq:4}
\ell_{\total} = \ell_{\gen} + \alpha\cdot \ell_{\type} + \beta\cdot \ell_{\connec}
\end{equation*}
where $\ell_{\gen}$ is the cross-entropy loss of the generation task, and $\ell_{\type}$ and $\ell_{\connec}$ are the cross-entropy losses of the discourse type and connective phrase predictions, respectively, with scalar weights $\alpha$ and $\beta$. We tuned $\alpha$ and $\beta$ on the \disco{} development sets, considering the values \{0.1, 0.5, 1\} for both weights. We then chose the best performing set of hyperparameter according to the higher \EXACT{} score on the appropriate development set. In all cases the resulting values were 0.1 for both weights.

The auxiliary heads of \AuxBertTrans{} and \AugAuxBertTrans{} also share the same architecture and hyper-parameters. For each auxiliary classifier we used one fully-connected layer, where the input dimension is 768, derived from \emph{BERT's pooler} output, and the output dimension is determined by the auxiliary output dimension (71 discourse markers and 12 discourse phenomena).   

We use the best-performing architecture and hyper-parameters specified by \citet{malmi2019encode} for the  \LaserTagger model. Specifically, we use the auto-regressive model, \texttt{AR-LaserTagger}, with an initialized BERT-Base Cased encoder and a small randomly initialized Transformer decoder. This model has shown better results on the fusion task compared to \texttt{FF-LaserTagger}, the non auto-regressive model.

\section{Experimental Details}
All experiments were performed on either one or two Nvidia GeForce GTX 1080 Ti GPUs, with two cores, 11 GB GPU memory per core, 6 CPU cores and 62.7 GB RAM.

We measured an average of ~8.5 hours for 45,000 training steps for \BertTrans{}, \AuxBertTrans{} and \AugAuxBertTrans{}, which is approximately a full `Wikipedia' epoch and about one-third of a full `Sports' epoch. To achieve full convergence, each model requires about 675K-900K training steps.

In \Cref{tab:dev-results} we provide the corresponding single-reference \EXACT{} validation performance for each reported test result. Notice that domain adaptation setups are not included within this table, since in such setups we use development data from the source domain.

\section{URLs of Code and Data}
As noted in \S \ref{sec:settings}, we provide here the URLs for the datasets and code we have used:  

\begin{itemize}
    \item \disco{} \citep{geva2019discofuse} A large scale dataset for sentence fusion:  \url{https://github.com/google-research-datasets/discofuse}
    \item Code and pre-trained weights of the pre-trained \emph{BERT-Base, Uncased} \citep{devlin2018bert} model:  
    \url{https://github.com/google-research/bert}
    \item Code for \LaserTagger{} \citep{malmi2019encode}: 
    \url{https://github.com/google-research/lasertagger}
    \item Code for BERTScore
    \citep{bert-score}:
    \url{https://github.com/huggingface/nlp/metrics/bertscore}
\end{itemize}

\end{document}